\pdfoutput=1

\documentclass[11pt]{article}

\usepackage[]{acl}

\usepackage{times}
\usepackage{latexsym}

\usepackage[T1]{fontenc}

\usepackage[utf8]{inputenc}

\usepackage{microtype}

\usepackage{inconsolata}

\usepackage{tcolorbox}
\usepackage{todonotes}
\usepackage{comment}

\usepackage{subfigure}
\usepackage{booktabs}
\usepackage{multirow}
\usepackage{pgfplots}
\usepackage{xcolor}
\usepackage{float}

\usepackage{filecontents}
\usetikzlibrary{pgfplots.statistics, pgfplots.colorbrewer} 
\usepackage{pgfplotstable}
\usepgfplotslibrary{colorbrewer}
\usepgfplotslibrary{groupplots}

\definecolor{tartunlp_violet}{HTML}{7268D8}
\definecolor{tartunlp_red}{HTML}{EF6650}
\definecolor{tartunlp_yellow}{HTML}{E0B12B}
\definecolor{tartunlp_black}{HTML}{282828}
\definecolor{tartunlp_green}{HTML}{4DB6AC}
\definecolor{tartunlp_blue}{HTML}{3185FF}
\definecolor{tartunlp_gray}{HTML}{9B9B9B}

\title{Teaching Llama a New Language Through Cross-Lingual Knowledge Transfer}

\author{Hele-Andra Kuulmets\textsuperscript{*} ~~~ Taido Purason\textsuperscript{*} ~~~ Agnes Luhtaru ~~~ Mark Fishel \\
Institute of Computer Science \\
  University of Tartu \\
        {\tt \{hele-andra.kuulmets, taido.purason, agnes.luhtaru, mark.fisel\}@ut.ee}}

\begin{document}
\maketitle
\begin{abstract}

This paper explores cost-efficient methods to adapt pretrained Large Language Models (LLMs) to new lower-resource languages, with a specific focus on Estonian. Leveraging the Llama 2 model, we investigate the impact of combining cross-lingual instruction-tuning with additional monolingual pretraining. Our results demonstrate that even a relatively small amount of additional monolingual pretraining followed by cross-lingual instruction-tuning significantly enhances results on Estonian. Furthermore, we showcase cross-lingual knowledge transfer from high-quality English instructions to Estonian, resulting in improvements in commonsense reasoning and multi-turn conversation capabilities. Our best model, named \textsc{Llammas}, represents the first open-source instruction-following LLM for Estonian. Additionally, we publish Alpaca-est, the first general task instruction dataset for Estonia. These contributions mark the initial progress in the direction of developing open-source LLMs for Estonian.

\end{abstract}

\renewcommand{\thefootnote}{\fnsymbol{footnote}}
\footnotetext[1]{Equal contribution}
\def\thefootnote{\arabic{footnote}}

\section{Introduction}

Instruction-tuning is a method for aligning large language models (LLMs) with human preferences \citep{ouyang2022training, mishra-etal-2022-cross, wei2021finetuned}. However, the majority of instruction-tuning datasets and advancements focus on English. Moreover, to benefit from instruction tuning, a strong foundation model is needed but due to the extensive training training data required, such models are available only for a few languages.

To overcome the lack of a strong foundation model in the target language, one could try to elicit non-English abilities from English-centric LLMs through cross-lingual instruction-tuning. In this setup, instructions are given in both English and the target language, often including a translation task to directly stimulate the alignment \citep{ranaldi2023empowering, ranaldi-pucci-2023-english, zhu2023extrapolating}. While empirical evidence indicates benefits from incorporating translation-following demonstrations into the training dataset, the best training strategy and its effectiveness with monolingual pretraining remain unclear.

In this paper, we investigate these aspects in the context of creating an instruction-following model for Estonian. We focus on a low-resource scenario where only a relatively small amount of monolingual data is available. By utilizing a novel general task instruction dataset, Alpaca-est, we examine the impact of combining monolingual pretraining with cross-lingual instruction-tuning using both general and translation task instructions. Our experiments with Llama 2 \citep{touvron2023llama2} demonstrate the benefits of translation task instructions when no monolingual data is available for additional pretraining. However, monolingual pretraining greatly diminishes the importance of the translation task.

Furthermore, we showcase that supplementing our instruction-tuning dataset consisting of Alpaca \citep{alpaca} and Alpaca-est with high-quality English instructions and English conversations further enhances results on Estonian through cross-lingual knowledge transfer. This is reflected in improved commonsense reasoning and the ability to engage in multi-turn conversations despite no Estonian conversations used during training. As a result, we present \textsc{Llammas} - the first open-source instruction-following conversational LLM for Estonian that achieves competitive zero-shot performance on multiple tasks.

\section{Related Work}

\subsection{Instruction Tuning}

Instruction-tuning is a method for guiding pre-trained LLMs to follow natural language instructions \citep{ouyang2022training, mishra-etal-2022-cross, wei2021finetuned, sanh2021multitask, chung2022scaling, wang-etal-2022-super}. For that purpose, both human-written and synthetic instructions generated with LLMs have been shown to work remarkably well \citep{wang-etal-2022-super, wang-etal-2023-self-instruct}.
One of the prerequisites for instruction-tuning is the availability of a strong pretrained language model, which due to high training costs is the major limiting factor for many to contribute to the development of LLMs. Fortunately, over the last year, a few foundation models \citep{workshop2022bloom, touvron2023llama, touvron2023llama2, jiang2023mistral} have been publicly released which somewhat mitigates the issue. However, the models are mostly trained on English and perform poorly on other languages.

A common method of acquiring instruction data is using strong proprietary models such as GPT-4 for generating instructions \cite{alpaca, vicuna2023, selfinstruct}. However, \citet{gudibande2023false} have shown that models trained on these generated datasets learn to imitate the style of strong LLMs but not necessarily the factuality.

\subsection{Cross-lingual Instruction Tuning}

Cross-lingual instruction tuning is a training method where the model is simultaneously instruction-tuned on instructions in multiple languages. Its goal is to strengthen cross-lingual semantic alignment in LLMs to make them understand and generate texts in a selected target language. In practice, it is one of the most cost-efficient ways to create instruction-following models for languages where data-heavy pretraining is not possible.

The approach has been explored, for example, by \citet{zhu2023extrapolating} and \citet{ranaldi2023empowering} who both use original and translated versions of Alpaca \citep{alpaca} dataset. Moreover, they both report additional benefits from supplementing the general task instruction datasets with translation task instructions. However, their approaches differ in the size of translation datasets. \citet{zhu2023extrapolating} use datasets that sometimes contain around 10 times more translation task instructions than general task instructions. \citet{ranaldi2023empowering} employ a translation task instruction dataset that contains only 20K instructions. Additionally, while \citet{zhu2023extrapolating} report benefits from using English to target language translations, \citet{ranaldi2023empowering} demonstrated that using both translation directions together is better than translating to only one direction. \citet{zhang2023bayling} propose to combine the task of strengthening cross-lingual semantic alignment and instruction-tuning via a multi-turn translation task. \citet{zhang2023plug} utilize the capabilities of LLMs to comprehend and execute instructions in a high-resource language by using that high-resource language as a pivot language during response generation for the target language.

\subsection{Monolingual Continued Pretraining}

Another way to improve the ability of English-centric pretrained LLMs to understand and generate content in a target language is via continued pretraining on data in the target language. For example, \citet{cui2023efficient} continue pretraining LLaMA family models on a large-scale monolingual Chinese corpus before the instruction-tuning. \citet{xu2023paradigm} show that continued pretraining with even a relatively small monolingual dataset can significantly improve the results of the translation instruction task. Moreover, they show that after continued pretraining only a small amount of high-quality parallel data is required to reach competent translation.

\subsection{Multilingual Models}

To create models that can follow instructions across diverse languages, multilingual pretraining can be combined with multilingual instruction tuning. For instance, \citet{wei2023polylm} pretrain a multilingual language model and then employ multilingual general task instructions generated through a self-instruct paradigm \citep{selfinstruct}.

\citet{yong-etal-2023-bloom} investigate strategies for adapting the multilingual language model BLOOM to new languages under resource-constrained settings. They find that adapter-based fine-tuning proves to be more effective than continued pretraining. Moreover, they demonstrate the advantages of multilingual instruction tuning over target language instruction tuning. \citet{lin2024mala} continue pretraining Llama-2-7B with low-rank adaptation \citep{hu2022lora} to develop a multilingual language model capable of encompassing 534 languages, including Estonian.

\section{Training Data}

\subsection{General Task Instructions}

\subsubsection{Alpacas}

We combine the original Stanford Alpaca dataset \citep{alpaca} with an Estonian version of it which we create by ourselves. We refer to the combination of these two datasets as \textbf{Alpacas}.

\paragraph{Stanford Alpaca \citep{alpaca}}

A general task instruction dataset generated with Self-Instruct framework \citep{wang-etal-2023-self-instruct}. In our experiments we use the cleaned version\footnote{\label{foot:alpaca-cleaned}\href{https://github.com/gururise/AlpacaDataCleaned}{https://github.com/gururise/AlpacaDataCleaned}} that consists of filtered Alpaca \cite{alpaca} instructions and GPT-4-LLM \citep{peng2023instruction}.

\paragraph{Alpaca-est}
Due to a lack of general task instruction data in Estonian, we generate an Estonian version of Alpaca. Following \citet{alpaca}, we first randomly sample from a set of Estonian seed instructions and use an LLM to generate new instructions based on the examples. Using \texttt{gpt-3.5-turbo-0613}\footnote{\label{foot:openai-models}\href{https://platform.openai.com/docs/models}{https://platform.openai.com/docs/models}}, we generate a total of 52,006 instructions for Estonian. The seed instruction set consists of 90 translated examples from the original Alpaca seed set and 17 new instructions written by the authors. We make Alpaca-est publicly available\footnote{\href{https://github.com/TartuNLP/alpaca-est}{https://github.com/TartuNLP/alpaca-est}}.

\subsubsection{High-Quality General Task Instructions}

We supplement Alpacas with high-quality English instructions that are not obtained with synthetic data generation using OpenAI models. In our dataset creation, we take inspiration from \citet{wang2023far, ivison2023camels}. We use Open Assistant 1 \citep{köpf2023openassistant} multi-turn conversations, taking the top-scoring English-only path from each conversation tree. We also take 10,000 examples of both Chain-of-Thought and FLAN-2 mixtures \citep{chung2022scaling, longpre2023flan} used in \citet{ivison2023camels}. We refer to this high-quality mixture of data in short as \textsc{\textbf{HQI}}.

\subsection{Translation Task Instructions} \label{data-translation-task-instructions}
We create translation task instructions from relatively low-quality translation bitexts: CCMatrix \citep{schwenk-etal-2021-ccmatrix}, WikiMatrix \citep{schwenk-etal-2021-wikimatrix}, OpenSubtitles \citep{lison-tiedemann-2016-opensubtitles2016}, and Europarl \citep{tiedemann-2012-parallel}. We filter the data with OpusFilter \citep{aulamo-etal-2020-opusfilter} using long word, sentence length, source-target length-ratio, character score, language-ID, terminal punctuation, and non-zero numerals filters. 

We use a setup in which 75\% of instructions prompt translation from English to Estonian, and 25\% prompt translation in the opposite direction. The goal of including a small amount of Estonian-English is to maintain the quality of English generation. We refer to this translation task instructions dataset as \textsc{\textbf{TrTask}}.

We supplement the relatively low-quality \textsc{\textbf{TrTask}} dataset with high-quality parallel data from WMT18 dev set \citep{bojar-etal-2018-findings} and MTee \citep{tattar2022open} held-out validation dataset. We refer to it as \textsc{\textbf{HQTrTask}}. In \textsc{\textbf{HQTrTask}} WMT18 dev set is given in a document-level format with documents exceeding 900 tokens split into multiple parts. To convert the translation examples to instructions we utilize 32 English and 13 Estonian prompt templates as \citet{sanh2021multitask} has demonstrated the importance of using a diverse set of prompts.

\subsection{Pretraining Data}  \label{sec:pretraining-data}

For pretraining, we use a subset of Estonian and English data from CulturaX \citep{nguyen2023culturax} to make the base model more familiar with Estonian but not forget English. Although the data in CulturaX has already gone through an extensive cleaning pipeline, we expand it by only allowing Estonian data that comes from websites ending with either .ee, .org, or .net. The pretraining is done with up to 5B tokens. We sample the data so that 75\% of CulturaX training documents are in Estonian while the rest are in English, to prevent English knowledge forgetting.

\section{Experimental Setup}

\subsection{Base Model}

To obtain the base model, we continue pretraining Llama-2-7B \cite{touvron2023llama2} with the additional 5B tokens of pretraining data described in Section \ref{sec:pretraining-data}. We call the base model \textsc{Llammas-base}. We use packing for pretraining which means that the training examples are concatenated to fill the model context. The training setup and parameters are outlined in Appendix~\ref{sec:training-parameters}. We publish our training code\footnote{\href{https://github.com/TartuNLP/llammas}{https://github.com/TartuNLP/llammas}}.

\subsection{Instruction-tuned Models}

Models instruction-tuned only with Alpacas or translation task instructions use the Alpaca prompting format \citep{alpaca}. The models relying on high-quality instructions (\textbf{HQI} or \textsc{\textbf{HQTrTask}}) are trained as conversational models with conversation format following \citet[][see Table~\ref{tab:chat-format}]{wang2023far}.

During the training, we calculate the loss only on responses, ignoring user input (including multi-turn) and instructions. The models are trained for 3 epochs. We picked the best checkpoint according to the validation loss, which was always the first checkpoint (trained for 1 epoch) in our experiments. See Appendix~\ref{sec:training-parameters} for other training details.

\begin{figure}
\begin{tikzpicture}
\pgfplotstableread[col sep=space]{continued_pretraining_est.tsv}\csvdata

	\pgfplotstabletranspose\datatransposed{\csvdata} 
    \begin{axis}[
        width  = 8cm,
        height = 5cm,
        major x tick style = transparent,
        x axis line style = {opacity=0},
        y axis line style = {opacity=0},
        axis y line = left,
        ylabel near ticks,
        ymajorgrids = true,
        ytick style={draw=none},
        ylabel = {Score},
        xticklabels = {
            CSR,
            QA,
            MT\textsubscript{EN-ET},
            MT\textsubscript{ET-EN},
            GEC,
        },
        xtick = data,
        /pgf/bar width={5pt},  
        ybar,
        xticklabel style = {align=center, font=\tiny},
        enlarge x limits=0.15,
        ymin=0, ymax=1,
        legend cell align=left,
        label style={font=\small },
        legend style={
                at={(1,1.2)},
                font=\small,
                anchor=north east,
                column sep=0.5ex,
                draw=none,
                legend columns = -1
        },
        legend style={/tikz/every even column/.append style={column sep=0.5cm}}]
    ]
        \addplot[style={tartunlp_red, fill=tartunlp_red,mark=none}]
             table[y index=1] {\datatransposed};

        \addplot[style={tartunlp_yellow,fill=tartunlp_yellow,mark=none}]
             table[y index=2] {\datatransposed};

        \addplot[style={tartunlp_violet,fill=tartunlp_violet,mark=none}]
             table[y index=3] {\datatransposed};
    
        \legend{0, 1B, 5B}
    \end{axis}
\end{tikzpicture}
    \caption{Results on Estonian tasks after fine-tuning Llama-2-7B with cross-lingual instruction-tuning dataset Alpacas. The colors of the bars indicate the size of the pretraining dataset.}
    \label{fig:pretraining-data-effect}
\end{figure}
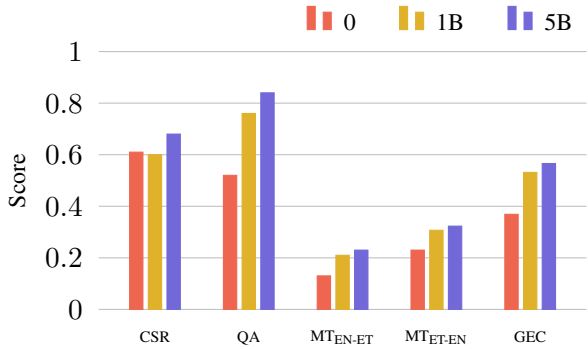

\subsection{Evaluation Datasets}
Following \citet{ranaldi2023empowering, zhu2023extrapolating}, we use EstQA \citep{kaver2021estqa}, an Estonian version of SQUAD \citep{rajpurkar2016squad} as one of the evaluation datasets. Since the original EstQA does not include a validation split, we create one ourselves by separating a small subset of training data for that purpose.  

We also evaluate our models on Estonian commonsense reasoning (CSR) and grammatical error correction (GEC) tasks. For commonsense reasoning, we use EstCOPA \citep{kuulmets2022estcopa}, which is an Estonian version of the COPA task \cite{roemmele2011choice}. EstCOPA includes both machine-translated and manually post-edited versions of COPA. We use the latter for our evaluations. Grammatical error correction is evaluated with EstGEC-L2 dataset\footnote{\label{foot:estgec}\href{https://github.com/tlu-dt-nlp/EstGEC-L2-Corpus}{https://github.com/tlu-dt-nlp/EstGEC-L2-Corpus}}. 

Finally, results for English-Estonian and Estonian-English translation (MT) tasks are reported using FLORES-200 devtest \citep{nllb2022}. It is important to note that, depending on the model, the translation task may be included into the training process, while the models are never exposed to any other evaluation tasks.

\subsection{Perfomance on English}

Ideally, our model should also perform reasonably well in English. If that was not the case it would mean that we might have washed out the pre-existing knowledge from the models. That could happen, for example, with overly extensive training on task-specific datasets. Naturally, it would be an indication that the model is not using its knowledge in English to generate answers in Estonian. To verify that our models can still understand English, we evaluate our best models on COPA, on an English subset of XQuAD \citep{artetxe2020cross}, and an English grammatical error correction task using the W\&I+LOCNESS test set \cite{bryant-etal-2019-bea}.

\begin{figure}
\begin{tikzpicture}
\pgfplotstableread[col sep=space]{continued_pretraining_trstep_gains.tsv}\csvdata

	\pgfplotstabletranspose\datatransposed{\csvdata} 
    \begin{axis}[
        width  = 8cm,
        height = 5cm,
        major x tick style = transparent,
        x axis line style = {opacity=0},
        y axis line style = {opacity=0},
        ytick style={draw=none},
        axis y line = left,
        ylabel near ticks,
        ymajorgrids = true,
        minor y tick num=1,
        yminorgrids = true,
        ylabel = {$\Delta_{Score}$},
        xticklabels = {
            CSR,
            QA,
            MT\textsubscript{EN-ET},
            MT\textsubscript{ET-EN},
            GEC,
        },
        y label style={at={(axis description cs:-0.1,.5)},anchor=south},
        xtick = data,
        /pgf/bar width={5pt},  
        ybar,
        xticklabel style = {align=center, font=\tiny},
        enlarge x limits=0.1,
        ymin=-0.15, ymax=0.15,
        legend cell align=left,
        label style={font=\small },
        legend style={
                at={(1,1.2)},
                font=\small,
                anchor=north east,
                column sep=0.5ex,
                draw=none,
                legend columns = -1
        },
        legend style={/tikz/every even column/.append style={column sep=0.5cm}}]
    ]
        \addplot[style={tartunlp_red, fill=tartunlp_red,mark=none}]
             table[y index=1] {\datatransposed};

        \addplot[style={tartunlp_yellow,fill=tartunlp_yellow,mark=none}]
             table[y index=2] {\datatransposed};

        \addplot[style={tartunlp_violet,fill=tartunlp_violet,mark=none}]
             table[y index=3] {\datatransposed};

        \legend{0, 1B, 5B}
    \end{axis}
\end{tikzpicture}
    \caption{Performance gained or lost on Estonian tasks after fine-tuning Llama-2-7B first on translation task and then on Alpacas compared to when translation task is omitted (Figure \ref{fig:pretraining-data-effect}). The colors of the bars indicate the size of the pretraining dataset.}
    
    \label{fig:pretraining-data-effect-with-translation}
\end{figure}
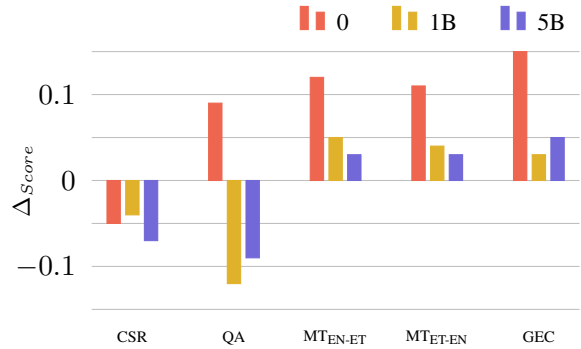

\begin{table*}[h!]\centering
\begin{small}
\begin{tabular}{llrrrrrrr}\toprule
\multicolumn{2}{l}{\multirow{2}{*}{Model}}&CSR &\multicolumn{2}{c}{QA} &MT\textsubscript{EN-ET} &MT\textsubscript{ET-EN} &GEC \\\cmidrule{4-5}
& &acc. &F1 &acc. &BLEU &BLEU &F0.5 \\ \midrule
\multicolumn{8}{l}{\textsc{Llammas-base} fine-tuned} \\\midrule
\textbf{(1)}&Alpacas &63.6 &46.5 &81 &22.5 &32.3 &56.6 \\
\textbf{(2)}&1) \textsc{TrTask} 2) Alpacas &59.2 &46.1 &73 &25.0 &34.5 &59.4 \\\midrule
\textbf{(3)} &Alpacas + \textsc{HQI} &\textbf{66.4} &52.9 &82 &23.1 &32.4 &59.4 \\
\textbf{(4)} &Alpacas + \textsc{HQI} + \textsc{HQTrTask} &\textbf{66.4} &\textbf{54.8} &\textbf{84} &22.6 &34.6 &60.3 \\
\textbf{(5)} &1) \textsc{TrTask} 2) \textbf{(4)} &62.2 &43.5 &76 &\textbf{26.9} &\textbf{36.9} &\textbf{61.2} \\ \midrule
\multicolumn{8}{l}{Commercial baselines} \\
\midrule
&GPT3.5-turbo &86.0 &34.2 &93 &26.0 &37.5 &63.4 \\
&GPT4 &98.4 &35.1 &97 &28.5 &37.7 &67.4 \\
\bottomrule
\end{tabular}
\end{small}
\caption{Results on Estonian tasks after fine-tuning \textsc{Llammas-base} on different cross-lingual instruction datasets. We call \textbf{(4)} \textsc{Llammas} and \textbf{(5)} \textsc{Llammas-mt}.}\label{tab:main-results}
\end{table*}

\subsection{Evaluation Metrics}

To evaluate commonsense reasoning and question-answering we use the assessments of GPT-4 Turbo\footref{foot:openai-models}. More precisely, we employ LLM-as-a-Judge \citep{zheng2023judging} with reference-guided grading where the model is asked to assess the correctness of the predicted answer given the reference answer and the task itself. We modified the evaluation prompt from \citet{zheng2023judging} to align with our tasks. We chose GPT-4 Turbo as the evaluator over ChatGPT\footref{foot:openai-models} to ensure the reliability of the results, as it demonstrated a significant improvement in assessment quality (specifically, a reduction in false positives) in our preliminary experiments. To reduce API usage costs, we base our QA accuracy report on 100 randomly chosen samples from the corresponding datasets and splits. When evaluating the commonsense reasoning task, we feed to GPT-4 Turbo only answers that we were not able to classify with a simple string comparison.  

We also report standard metrics for most of the tasks. For question answering and grammatical error correction we report F1 and M2 scorer\footnote{\href{https://github.com/TartuNLP/estgec/tree/main/M2_scorer_est}{https://github.com/TartuNLP/estgec/tree/main/M2\_scorer\_est}} \cite{dahlmeier-ng-2012-better} or ERRANT \cite{bryant-etal-2017-automatic} F0.5, respectively. For translation tasks we calculate BLEU\footnote{sacreBLEU signature: \texttt{nrefs:1|case:mixed|eff:no|tok:13a| smooth:exp|version:2.3.1}} \citep{papineni-etal-2002-bleu} and chrF++\footnote{sacreBLEU signature: \texttt{nrefs:1|case:mixed| eff:yes|nc:6|nw:2|space:no|version:2.3.1}} \citep{popovic-2017-chrf} using sacreBLEU \citep{post-2018-call}, and COMET \citep{rei-etal-2020-comet} scores using the \texttt{unbabel-wmt22-comet-da} model \citep{rei-etal-2022-comet}.

\subsection{Evaluation Prompts}

During the development phase, the performance on EstCOPA, EstQA, and their English equivalents is measured with 8 different prompts. The English prompts are from \citet{wei2021finetuned}, while prompts for Estonian tasks are written by the authors. On development datasets, we report the best score across the 8 prompts, while on test datasets, we only report the scores obtained with the best prompt according to the development datasets. For machine translation and grammatical error correction tasks, we use the same single prompt during the development and test phases (see Table~\ref{tab:eval-prompts}).

\begin{table*}[h]\centering
\begin{small}
\begin{tabular}{lrrrrrrrrr}\toprule
\multirow{2}{*}{Model}& \multirow{2}{*}{Param.} &\multicolumn{3}{c}{ET$\rightarrow$EN} & &\multicolumn{3}{c}{EN$\rightarrow$ET} \\\cmidrule{3-5}\cmidrule{7-9}
&&BLEU &chrfF++ &COMET & &BLEU &chrfF++ &COMET \\\midrule
MTee \citep{tattar2022open}& 227M &36.7 &61.3 &88.5 & &27.6 &56.9 &89.2 \\
NLLB-MoE \citep{nllb2022} & 54.5B &\textbf{38.8} &62.6 &89.3 & &27.1 &56.1 &91.4 \\\midrule
GPT-3.5-turbo&- &37.5 &63.0 &89.5 & &26.0 &56.3 &91.7 \\
GPT-4-turbo&- &37.7 &\textbf{63.8} &\textbf{89.7} & &\textbf{28.5} &\textbf{58.4} &\textbf{92.6} \\\midrule
\textsc{Llammas} (ours)& 7B &34.6 &59.2 &89.0 & &22.6 &51.8 &91.0 \\
\textsc{Llammas-mt} (ours) & 7B &36.9 &61.2 &89.1 & &26.9 &56.4 &91.9 \\
\bottomrule
\end{tabular}
\end{small}
\caption{Translation metric scores on FLORES-200 devtest \citep{nllb2022}.}\label{tab:translation}
\end{table*}

\begin{table*}[!htp]\centering
\begin{small}
\begin{tabular}{lrrrrrrr}\toprule
\multicolumn{1}{l}{\multirow{2}{*}{Model}} & \multicolumn{3}{c}{ET} && \multicolumn{3}{c}{EN} \\\cmidrule{2-4}\cmidrule{6-8}
& P & R & F$_{0.5}$ && P & R & F$_{0.5}$ \\\midrule

GPT-3.5-turbo & 69.6 & 46.7 & 63.4 && 53.6 & 70.1 & 56.3\\
GPT-4 & \textbf{74.3 }& \textbf{49.2} & \textbf{67.4} && 56.7 & \textbf{71.6} & \textbf{59.1 }\\
\midrule
\textsc{Llammas} (ours)  &67.6 &42.2 & 60.3 && \textbf{58.0} & 59.5 & 58.3 \\
\textsc{Llammas-mt} (ours) & 68.0 & 43.6 & 61.2 && 55.9 & 59.3 & 56.6 \\
\bottomrule
\end{tabular}
\end{small}
\caption{GEC scores on EstGEC-L2 and W\&I+LOCNESS test sets.}\label{tab: gec-results}
\end{table*}

\section{Experiments and Results}

Our experiments are divided into two main sections. In the first section, we pretrain Llama-2-7B on different amounts of pretraining data and investigate the effect of it on cross-lingual instruction-tuning that is done with translation task and general task instructions (Alpacas).

In the second section, we study the influence of supplementing Alpacas with high-quality English instructions, translations, and conversations to the results on Estonian.

\subsection{Continued Pretraining of Llama 2}

We compare three base models. First, Llama-2-7B without any additional pretraining. Second, the checkpoint of \textsc{Llammas-base} that has seen 1B tokens of pretraining data. Third, \textsc{Llammas-base} trained on the entire pretraining dataset of 5B tokens. We instruction-tune all three models on Alpacas that consisting of Estonian and English general task instructions. The results of the three models are compared in Figure \ref{fig:pretraining-data-effect}. We observe performance gains on all Estonian tasks as the size of the pretraining dataset increases.

In our preliminary experiment (included into the ablation study, Section {\ref{first_experiment}}) we observed that after additional pretraining of Llama-2-7B with 1B tokens the benefits of using translation task during fine-tuning diminished. To assess whether this trend persists with even larger pretraining, we instruction-tune the base models with a dataset that consists of both translation and general task instructions, i.e., \textsc{\textbf{TrTask}} and Alpacas. We adopt sequential training based on our preliminary experiment (Section \ref{first_experiment}), which indicated that this setup has a milder negative impact on performance in zero-shot tasks.

Figure \ref{fig:pretraining-data-effect-with-translation} shows the performance gained or lost for each task and base model with the translation task used as the first step during instruction-tuning. We can see that without additional pretraining, the translation task significantly improves the results for QA, machine translation, and GEC. However, the benefit diminishes greatly when the pretraining step is introduced. For QA and commonsense reasoning, omitting the translation task after pretraining tends to produce stronger results compared to models where pretraining is followed by the translation task.
 
\subsection{Beyond Alpacas: Knowledge Transfer via High-Quality English Instructions}

Instruction-tuning datasets generated with Self-Instruct \citep{wang-etal-2023-self-instruct} might suffer from various issues that lower the overall quality of the dataset\footref{foot:alpaca-cleaned}. Meanwhile, it has been shown that it is possible to achieve remarkably strong performance with just 1,000 high-quality training examples \citep{zhou2023lima}. In light of this, we hypothesize that supplementing the Alpacas dataset with a set of high-quality instructions could improve the models. However, as there are no high-quality instruction datasets available for Estonian, we use only high-quality English instructions (\textbf{HQI}). For comparison, we train a model where high-quality English instructions are supplemented with high-quality translation task instructions (\textbf{\textsc{HQTrTask}}).

The results are shown in Table \ref{tab:main-results}. Compared to the baseline model (1) that is trained on just Alpacas, we observe a somewhat surprising increase in all scores when Alpacas is supplemented with high-quality English instructions (model (3)). This suggests that there is a positive cross-lingual knowledge transfer from the added high-quality English instructions into Estonian. Moreover, combining high-quality English instructions with high-quality translation tasks further enhances the knowledge transfer (model (4)). We call this model \textsc{Llammas}. However, we observe that the best results for EN$\rightarrow$ET, ET$\rightarrow$EN, and GEC are obtained with a model that is trained sequentially, with \textsc{HQTrTask} as the first step of fine-tuning (model (5)). We call this model \textsc{Llammas-mt}.

Models (3) -- (5) are trained with the data in chat format (see Table~\ref{tab:chat-format}), since \textsc{HQI} contains English conversational data from Open Assistant~1. Through manual evaluation with 5 conversations (up to 6 turns), we determine that model (4) (\textsc{Llammas}) can adequately engage in multi-turn conversations. It can recall content from previous turns and respond to user requests fairly well. However, we also see that the model sometimes makes grammatical mistakes and uses words or phrases that a native Estonian speaker would not use. Many of these phrases sound like translations from English. An example conversation can be seen in Table~\ref{tab:llammas-conversation-example}. The model's conversational ability suggests that the model has learned to hold a multi-turn conversation in Estonian through cross-lingual transfer, however, more experiments would be needed to confirm that.

\subsection{Results on Translation Task}

Conventional neural machine translation (NMT) models leverage tens of millions of parallel sentences along with the use of monolingual corpora. In contrast, \textsc{Llammas-mt} uses a modest 1~million sentence pairs from relatively low-quality parallel data sources and a small number of sentences from high-quality sources. In combination with general task instructions, this results in a competitive translation model, as presented in Table~\ref{tab:translation}. We can see that \textsc{Llammas-mt} outperforms  \textsc{Llammas} although, in terms of COMET, which is more highly correlated with human judgments \citep{freitag-etal-2022-results}, \textsc{Llammas} still seems competitive.

When comparing \textsc{Llamas-translate} to the open-source encoder-decoder models MTee and NLLB-MoE, \textsc{Llamas-translate} achieves better scores on COMET and similar scores on BLEU and chrF++. On ET$\rightarrow$EN \textsc{Llammas-mt} is outperformed by NLLB-MoE, however, it outperforms MTee on COMET and achieves a similar score in chrF++. We can also see that \textsc{Llammas-mt} is competitive with GPT-3.5-turbo, however it is outperformed by GPT-4-turbo (for used prompt, see Figure~\ref{tab:openai-mt-prompt}).

\subsection{Results on Grammatical Error Correction}

LLMs are good at text correction, yet they frequently make extensive edits that diverge from traditional GEC metrics, known for preferring minimal modifications \cite{coyne2023analyzing}. This tendency is apparent in English, where the models exhibit higher recall than precision (see Table \ref{tab: gec-results}). For Estonian, in contrast, the models show higher precision but reduced recall, indicating a different correction pattern from Estonian. We leave further exploration of that phenomenon for future work. Finally, we can see that translation task instructions (\textsc{TrTask}, used for training \textsc{Llammas-mt}) enhance performance in Estonian which is in accordance with our earlier experiments.

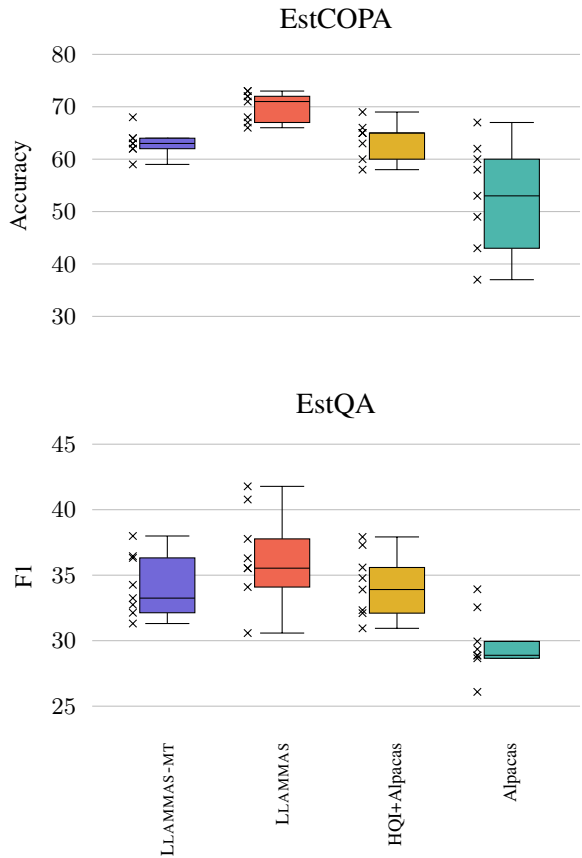
\begin{figure}[t]
\centering
\begin{tikzpicture}
	\pgfplotstableread[col sep=space]{combined_robustness.csv}\csvdata
	\pgfplotstabletranspose\datatransposed{\csvdata} 
	\begin{groupplot}[group style={group size= 1 by 2},height=5.75cm,width=8cm]

            \nextgroupplot[
            title=EstCOPA,
            title style={yshift=-2ex},
		boxplot/draw direction = y,
		x axis line style = {opacity=0},
		axis x line* = bottom,
		axis y line = left,
		enlarge y limits,
            cycle list={},
		ymajorgrids,
            ymin = 30, ymax=80,
            ytick distance=10,
		xtick = {1, 2, 3, 4},
            ylabel near ticks,
            boxplot/box extend=0.48,
		xticklabel style = {align=center, font=\tiny, rotate=90},
		xticklabels = {Llammas-mt, HQI + HQT + Alpacas, HQI + Alpacas, Alpacas},
            xticklabels={,,,,},
            yticklabel style = {font=\small},
		xtick style = {draw=none},
		ylabel = {Accuracy},
            label style={font=\small},
            width=8cm,
            height=5.75cm,
            xlabel near ticks,
            y axis line style = {opacity=0},
            ytick style={draw=none}
	]
            
    \addplot+[boxplot, fill=tartunlp_violet, draw=black] table[y index=5] {\datatransposed};
    \addplot[only marks, mark=x, mark color=tartunlp_gray] table[y index=5, x expr={1-0.3}] {\datatransposed};
    
    \addplot+[boxplot, fill=tartunlp_red, draw=black] table[y index=6] {\datatransposed};
    \addplot[only marks, mark=x, mark color=tartunlp_gray] table[y index=6, x expr={2-0.3}] {\datatransposed};
    
    \addplot+[boxplot, fill=tartunlp_yellow, draw=black] table[y index=7] {\datatransposed};
    \addplot[only marks, mark=x, mark color=tartunlp_gray] table[y index=7, x expr={3-0.3}] {\datatransposed};
    
    \addplot+[boxplot, fill=tartunlp_green, draw=black] table[y index=8] {\datatransposed};
    \addplot[only marks, mark=x, mark color=tartunlp_gray] table[y index=8, x expr={4-0.3}] {\datatransposed};

        \nextgroupplot[
            title=EstQA,
            title style={yshift=-2ex},
		boxplot/draw direction = y,
		x axis line style = {opacity=0},
		axis x line* = bottom,
		axis y line = left,
		enlarge y limits,
            cycle list={},
		ymajorgrids,
            ymin = 25, ymax=45,
		xtick = {1, 2, 3, 4},
            ylabel near ticks,
            boxplot/box extend=0.48,
            yticklabel style = {font=\small},
		xticklabel style = {align=center, font=\scriptsize, rotate=90},
		xticklabels = {\textsc{Llammas-mt}, \textsc{Llammas}, HQI+Alpacas, Alpacas},
		xtick style = {draw=none}, 
		ylabel = {F1},
            label style={font=\small },
            width=8cm,
            height=5.75cm,
            ytick distance=5,
            y axis line style = {opacity=0},
            ytick style={draw=none},
	]
            
    \addplot+[boxplot, fill=tartunlp_violet, draw=black] table[y index=1] {\datatransposed};
    \addplot[only marks, mark=x, mark color=tartunlp_gray] table[y index=1, x expr={1-0.3}] {\datatransposed};
    
    \addplot+[boxplot, fill=tartunlp_red, draw=black] table[y index=2] {\datatransposed};
    \addplot[only marks, mark=x, mark color=tartunlp_gray] table[y index=2, x expr={2-0.3}] {\datatransposed};
    
    \addplot+[boxplot, fill=tartunlp_yellow, draw=black] table[y index=3] {\datatransposed};
    \addplot[only marks, mark=x, mark color=tartunlp_gray] table[y index=3, x expr={3-0.3}] {\datatransposed};
    
    \addplot+[boxplot, fill=tartunlp_green, draw=black] table[y index=4] {\datatransposed};
    \addplot[only marks, mark=x, mark color=tartunlp_gray] table[y index=4, x expr={4-0.3}] {\datatransposed};

\end{groupplot}
 
\end{tikzpicture}
\caption{EstCOPA development set accuracy and EstQA development set F1-score of 8 prompts on models fine-tuned from \textsc{Llammas-base} (see Table~\ref{tab:main-results}).}
\label{fig:robustness-combined}
\end{figure}

\subsection{Results on XQUAD and COPA}

The results on English QA and commonsense reasoning tasks are shown in Table \ref{tab:english-results}. On the QA task, \textsc{Llammas} achieves similar accuracy in English and Estonian (83\% vs 84\%). However, we observed that \textsc{Llammas} is more chatty in English, resulting in longer answers and therefore lower F1 score when compared to  Estonian.
Finally, we observe that \textsc{Llammas} solves commonsense reasoning problems significantly better in English than in Estonian (80.6\% vs 66.4\%) This indicates that \textsc{Llammas} is still not able to utilize all the reasoning capabilities it has in English when the input is given in Estonian.\footnote{Hence the name \textsc{Llammas} as in Estonian the word \textit{lammas} means \textit{sheep}.}

\subsection{Robustness on Diverse Prompts}
We look into the distribution of metric scores on 8 development prompts (Table~\ref{tab:main-results}) to assess the robustness of our models when encountering various input prompts.

EstCOPA shows an increase in robustness and average scores with various prompts when high-quality English instructions are used (see Figure~\ref{fig:robustness-combined}). This is even further increased by the addition of high-quality translation instructions. While having lower scores than the models without a translation step, \textsc{Llammas-mt} still displays good robustness. On EstQA, however, we don't see the same trend. There is an increase in the median of the metric score, yet the robustness does not increase. For models involving the use of high-quality data, the lowest-scoring prompts still achieve higher F1 scores than the median of the model fine-tuned on Alpacas.

\begin{table*}[h]\centering
\small
\begin{tabular}{lrrrrrr}\toprule
Model &MMLU &TruthfulQA &WinoGrande &TriviaQA &HellaSwag \\
\midrule
Llama-2-7B  &45.8 &32.1 &68.8 &52.4 &76.0 \\
\textsc{Llammas-base}  &42.6 &31.9 &70.0 &49.8 &74.8 \\
\midrule
difference &-3.1 &-0.2 &1.2 &-2.6 &-1.2 \\
\bottomrule
\end{tabular}
\caption{Comparing the performance of Llama-2-7B and \textsc{Llammas-base} on different English benchmarks.}\label{tab:extended_eval}
\end{table*}

\begin{table}[h]\centering
\begin{small}
\begin{tabular}{lrrrr}\toprule
{Model}&CSR &\multicolumn{2}{c}{QA} \\
&acc. &F1 &acc. \\
\midrule
Alpacas &63.4 &30.4 &85 \\
1) TrTask 2) Alpacas &70.2 &29.5 &81 \\ \midrule
Alpacas + HQI &78.6 &33.3 &\textbf{87} \\
\textsc{Llammas} &\textbf{80.6} &\textbf{41.0}  &83 \\
\textsc{Llammas-mt} &73.6 &31.4 &82 \\\midrule
GPT3.5 &95.2 &30.7 &95 \\
GPT4 &99.8 &33.2 &96 \\
\bottomrule
\end{tabular}
\end{small}
\caption{Results on English commonsense reasoning and question answering.} \label{tab:english-results}
\end{table}

\subsection{Extended Evaluation on English}

To better understand how the performance on English is affected by continued pretraining on data that mostly contains Estonian, we extend our evaluation beyond English datasets for which Estonian equivalents are available. More precisely, we conduct additional evaluation on 5 popular English benchmarks. The benchmarks are MMLU \citep{hendryckstest2021} which covers 57 tasks with different levels of difficulties; TruthfulQA \citep{lin-etal-2022-truthfulqa} that measures truthfulness with questions designed to cause imitative falsehoods; WinoGrande \citep{10.1145/3474381}, a pronoun resolution challenge; TriviaQA \citep{joshi-etal-2017-triviaqa}, a question answering dataset; and HellaSwag \citep{zellers-etal-2019-hellaswag}, a commonsense reasoning task. We use lm-evaluation-harness \citep{eval-harness} and compare Llama-2-7B and \textsc{Llammas-base} on these benchmarks. We report 5-shot accuracy on MMLU and 0-shot on other benchmarks. 

\begin{table}[h]\centering
\small
\begin{tabular}{lrrrr}\toprule
Model &Hum. &STEM &Social &Other \\
\midrule
Llama-2-7B &43.3 &37.0 &51.5 &52.7  \\
\textsc{Llammas-base} &40.0 &34.7 &47.7 &49.7  \\
\midrule
difference &-3.3 &-2.3 &-3.8 &-3.0 \\
\bottomrule
\end{tabular}
\caption{Performance of Llama-2-7B and \textsc{Llammas-base} across categories in MMLU benchmark.}\label{tab:extended_eval_MMLU}
\end{table}

The results in Table \ref{tab:extended_eval} show that \textsc{Llammas-base} drops only slightly in performance on 4 out of 5 English benchmarks. On average, the difference is 1.2\%. The biggest contributor to the difference is MMLU from which tasks covering humanities and social sciences have the weakest accuracy when compared to Llama-2-7B (Table \ref{tab:extended_eval_MMLU}). Overall, we notice that the difference is larger for benchmarks that measure world knowledge (MMLU, TriviaQA) and smaller for commonsense reasoning tasks (winogrande, HellaSwag). The least affected by continued pretraining is TruthfulQA.

\section{Ablation Study}

\subsection{Instruction-Tuning: Sequentially or with a Combined Dataset?} \label{first_experiment}

Previous research has explored approaches that combine translation and general task instructions for cross-lingual instruction-tuning \citep{ranaldi-pucci-2023-english, ranaldi2023empowering, zhu2023extrapolating}. However, these approaches combine both types of instructions into a single dataset for model fine-tuning. We hypothesize that such setup, especially when a significantly larger translation task dataset is used (e.g. by \citeauthor{zhu2023extrapolating}, \citeyear{zhu2023extrapolating}), may diminish the contribution of general task instructions during the training, adversely impacting the model's ability to generalize to new tasks.

To test the hypothesis we compare fine-tuning Llama-2-7B on a combined dataset to fine-tuning it with sequential training. The latter involves first training the model on the translation task and then on general task instructions. We replicate the experiment with Llama-2-7B further pretrained on 1B tokens, to validate the consistency of results when the pretraining step is included. We use context size of 224 and, following \citet{zhu2023extrapolating}, only English to target language translations (\textbf{\textsc{TrTask}\textsubscript{EN$\rightarrow$ET}}). We compare the results with baselines where translation task data is entirely omitted.

The results in Table \ref{tab:translations-task-effect} show that fine-tuning Llama-2-7B on translation task improves most results (except commonsense reasoning). Combined training is particularly beneficial for EN$\rightarrow$ET and grammatical error correction. The latter aligns with the improvement in EN$\rightarrow$ET as MT and GEC are similar tasks and often approached in a similar way \citep{junczys-dowmunt-etal-2018-approaching}. However, QA and ET$\rightarrow$EN gain more from sequential training. It is especially notable for ET$\rightarrow$EN where general task instructions recover the performance after the initial degradation.

However, we observe that when pretraining Llama-2-7B on 1B tokens is included, the performance generally suffers when translation task instructions are used. Exceptions are English-Estonian and grammatical error correction that naturally benefit from the translation task.

Finally, we can see that EN$\rightarrow$ET is rather weak on pretrained Llama-2-7B after fine-tuning on just Alpacas. However, including the task drastically hurts the performance of ET$\rightarrow$EN translation task.

\subsection{Translation Data: The Impact of Quality and Quantity}
In Section \ref{first_experiment} we found that language-specific pretraining of Llama-2-7B followed by fine-tuning on just Alpacas outperforms the same base model fine-tuned on both translation and general task instructions. Combining the datasets (\textbf{\textsc{TrTask}\textsubscript{EN$\rightarrow$ET} + Alpacas}) yielded weaker scores than sequential training (\textbf{1) \textsc{TrTask}\textsubscript{EN$\rightarrow$ET} 2) Alpacas}). To address the potential negative influence from the imbalanced dataset, where translation instructions outnumber general task instructions by about 10 times, we conduct an experiment with a balanced dataset. We fine-tune the base model with a dataset combining general task instructions with 100K translation task instructions (similar in size to Alpacas) from the data mix described in Section \ref{data-translation-task-instructions}. Table \ref{tab:quantity-vs-quality} shows that the model does not outperform the Alpacas baseline. 

Additionally, we train the base model with a dataset combining general task instructions with a small set of high-quality translation task instructions from MTee held-out validation sets \citep{tattar2022open} and WMT18 development set \citep{bojar-etal-2018-findings}. This model also does not outperform the baseline model, except in GEC which seems to benefit from high-quality translation task.  

\subsection{Translation Data: Single Translation Direction or Both?}
We investigate the effect of EN$\rightarrow$ET : ET$\rightarrow$EN translation direction proportion in our data. From Table~\ref{tab:ablation-tr-direction}, we can see that for all tasks, having only EN$\rightarrow$ET direction is not optimal when translation data is used. For MT\textsubscript{ET$\rightarrow$EN} and GEC 25\% ET$\rightarrow$EN seems to offer the best scores, while for other tasks 50\% offers the highest scores. For CSR, having no translation data at all offers the highest accuracy.

\section{Conclusion}
We successfully adapt Llama 2 to Estonian by creating \textsc{Llammas} - an instruction-following model for Estonian. Additionally, we release Alpaca-est, an Alpaca-style general task instruction dataset for Estonian. Our work has shown competitive results for tasks such as question-answering, machine translation, and grammatical error correction in Estonian while keeping solid results for English. We have also identified signs of cross-lingual transfer from English to Estonian and investigated the effects of translation bitexts in the fine-tuning process. This work marks the first step towards open-source LLMs for Estonian.

\section*{Limitations}
The key limitation of this work is the dependence on data generated with OpenAI's proprietary LLMs. As \citet{gudibande2023false} have found, these generated datasets result in the imitation of the proprietary LLM's style but not necessarily factuality. 
Secondly, due to the limited number of benchmarks for Estonian, our evaluation is limited to a rather small number of NLP tasks. Because of the early stages of the research on capabilities and harmlessness, the model will be limited to research purposes.

\section*{Ethics}
We believe that extending open-source large language models to previously uncovered languages poses a net positive impact as it allows more people access to them. However, the currently released model lacks safety evaluation, meaning that it should be used only for research purposes. Furthermore, the self-instruct style generated instructions have not been manually checked, increasing the risks (for example bias) even more. Further research into evaluating the harmlessness and helpfulness of LLMs for Estonian is needed, as this has not been done for proprietary LLMs that support Estonian either.

\section*{Acknowledgements}

This work was supported by the Estonian Research Council grant PRG2006 (Language Technology for Low-Resource Finno-Ugric Languages and Dialects). All computations were performed on the LUMI Supercomputer through the University of Tartu’s HPC center.

\bibliography{anthology,custom}

\clearpage
\appendix

\section{Training Parameters}
\label{sec:training-parameters}
The context length in our training experiments is 1024 tokens with the overlapping examples truncated. The models are trained with \texttt{bf16} precision using DeepSpeed \cite{rasley2020deepspeed}. A learning rate of 2e-5 is used and is linearly decayed to 2e-6. During pretraining a batch size of 256 is used and during instruction-tuning the batch size is 128. We train our models on 4 AMD MI250x GPUs (acting as 8 GPUs) on the LUMI supercomputer.

The pretraining on 5B tokens took 1184 GPU-hours (\textsc{Llammas-base}). Instruction-tuning of \textsc{Llammas} took 80 GPU-hours (3 epochs). Instruction-tuning on translation data (\textsc{TrTask}) for \textsc{Llammas-mt} took 190 GPU-hours (3 epochs), in addition to the instruction-tuning on the general instructions (i.e, fine-tuning \textsc{Llammas}).

\section{Sizes of Datasets}

Training, test and validation dataset sizes are shown in Tables \ref{tab:train-dataset-sizes} and \ref{tab:validation-dataset-sizes}.

\setlength{\tabcolsep}{2pt}
\begin{table}[!htp]
\small
\begin{tabular}{lr}\toprule
\textbf{General task instructions} & \\
\midrule
Alpaca-cleaned \citep{alpaca} &52 000 \\
Alpaca-est (ours) &52 006 \\
\midrule
\textbf{HQI} & \\
\midrule
CoT \citep{chung2022scaling, ivison2023camels} &10 000 \\
FlanV2 \citep{chung2022scaling, ivison2023camels} &10 000 \\
Open Assistant 1 \citep{köpf2023openassistant} &2 363 \\
\bottomrule
\toprule
\textbf{Translation task instructions} & \\
\midrule
{\textbf{\textsc{TrTask}}} & \\
\midrule
CCMatrix \citep{schwenk-etal-2021-ccmatrix} &500 000 \\
WikiMatrix \citep{schwenk-etal-2021-wikimatrix} &400 000 \\
Europarl \citep{tiedemann-2012-parallel} &50 000 \\
OpenSubtitles \citep{lison-tiedemann-2016-opensubtitles2016} &50 000 \\
\midrule
{\textbf{\textsc{HQTrTask}}} & \\
\midrule
WMT18 dev (doc. level) \citep{bojar-etal-2018-findings} &245 \\
MTee valid held-out (general) \citep{tattar2022open} &1 528 \\
\midrule
{\textbf{Additional HQ translation data}} & \\
\midrule
MTee valid held-out (all) \citep{tattar2022open} &4 353 \\
WMT18 dev (sent. level) \citep{bojar-etal-2018-findings} &2 000 \\
\bottomrule
\end{tabular}
\caption{Sizes of instruction datasets (number of examples).}\label{tab:train-dataset-sizes}
\end{table}

\setlength{\tabcolsep}{3pt}
\begin{table}[!htp]\centering
\small
\begin{tabular}{lrr}\toprule
&Validation &Test \\
\midrule
\textbf{Question Answering} & & \\
\midrule
EstQA \citep{kaver2021estqa} &85 &603 \\
XQuAD \citep{artetxe2020cross} &1 190 &- \\
\midrule
\textbf{Commonsense Reasoning} & & \\
\midrule
EstCOPA \citep{kuulmets2022estcopa} &100 &500 \\
COPA \citep{roemmele2011choice} &100 &500 \\
\midrule
\textbf{Grammatical Error Correction} & & \\
\midrule
EstGEC-L2\footref{foot:estgec}  &879 &2 029 \\
W\&I+LOCNESS \cite{bryant-etal-2019-bea} &4 385 &4 477 \\
\midrule
\textbf{Machine Translation} & & \\
\midrule
FLORES-200 \citep{nllb2022} &997 &1 012 \\
\bottomrule
\end{tabular}
\caption{Sizes of evaluation and test datasets (number of examples). The entire XQUaD was used for both validation and testing.}\label{tab:validation-dataset-sizes}
\end{table}

\section{Ablation Study Tables}
\label{sec:ablation-study}

Results of ablation experiments are shown in Tables \ref{tab:translations-task-effect},  \ref{tab:quantity-vs-quality}, and \ref{tab:ablation-tr-direction}.

\section{Evaluation Prompts}
\label{sec:prompts}

Prompts for each evaluation task are shown in Figure \ref{tab:eval-prompts}. Alpaca instruction format is shown in Figure \ref{tab:alpaca-format} and chat format for training \textsc{Llammas} is shown in Figure \ref{tab:chat-format}. The prompt used for evaluating OpenAI models on MT task is shown in Figure \ref{tab:openai-mt-prompt}.

\section{Example Conversation with \textsc{Llammas}}

Table \ref{tab:llammas-conversation-example} shows an example multi-turn conversation with \textsc{Llammas} held in Estonian. 

\begin{table*}[h]\centering
\small
\begin{tabular}{lrrrrrr}\toprule
&CSR &QA &GEC &MT\textsubscript{EN$\rightarrow$ET} &MT\textsubscript{ET$\rightarrow$EN} \\
\textbf{} &acc. &acc. &F0.5 &BLEU &BLEU \\\midrule
\multicolumn{6}{c}{Llama-2-7B} \\\midrule
\textsc{TrTask}\textsubscript{EN$\rightarrow$ET} + Alpacas &\textbf{58} &61.2 &\underline{\textbf{55.1}} &\underline{\textbf{24.6}} &1.5 \\
1) \textsc{TrTask}\textsubscript{EN$\rightarrow$ET} 2) Alpacas &\textbf{58} &\underline{\textbf{64.7}} &51.2 &24.5 &\underline{\textbf{27.4}} \\\midrule
Alpacas &\underline{\textbf{61}} &51.8 &34.2 &13.9 &24.8 \\\midrule
\multicolumn{6}{c}{Llama-2-7B pretrained on 1B tokens of Estonian-centric data} \\\midrule
\textsc{TrTask}\textsubscript{EN$\rightarrow$ET} + Alpacas &53 &63.5 &\underline{\textbf{57.5}} &24.4 &1.5 \\
1) \textsc{TrTask}\textsubscript{EN$\rightarrow$ET} 2) Alpacas &\textbf{55} &\textbf{70.6} &55.5 &\underline{\textbf{25.7}} &\textbf{23.0} \\\midrule
Alpacas &\underline{\textbf{66}} &\underline{\textbf{74.1}} &50.5 &20.8 &\underline{\textbf{32.4}} \\
\bottomrule
\end{tabular}
\caption{Comparison of cross-lingual training strategies across two different base models. Results are reported on development datasets.} \label{tab:translations-task-effect}
\end{table*}

\begin{table*}[h]\centering
\begin{small}
\begin{tabular}{lcrrrrrr}\toprule
\multirow{2}{*}{Model} &TrTask&CSR &QA &MT\textsubscript{EN$\rightarrow$ET} &MT\textsubscript{ET$\rightarrow$EN} &GEC \\
\textbf{} &size &acc. &acc. &BLEU &BLEU &F0.5 \\\midrule
\textbf{TrTask\textsubscript{EN$\rightarrow$ET} + Alpacas} &1M &53 &63.5 &\textbf{24.4} &1.5 &\textbf{57.5} \\
\textbf{TrTask\textsubscript{EN$\rightarrow$ET} + Alpacas} &100K &56 &71.8 &21.1 &1.6 &56.2 \\
\textbf{TrTask\textsubscript{High quality EN$\rightarrow$ET} + Alpacas} &6K &57 &69.4 &22.2 &3.6 &\textbf{57.5} \\ \midrule
\textbf{Alpacas} & - &\textbf{66} &\textbf{74.1} &20.8 &\textbf{32.4} &50.5 \\
\bottomrule
\end{tabular}
\caption{Quantity vs quality: examining the impact of translation task dataset composition. Results are reported on development datasets.}\label{tab:quantity-vs-quality}
\end{small}
\end{table*}

\setlength{\tabcolsep}{6pt}
\begin{table*}[h!]\centering
\begin{small}
\begin{tabular}{lcrrrrrr}\toprule
\multirow{2}{*}{Model} &TrTask &CSR &QA &MT\textsubscript{EN$\rightarrow$ET} &MT\textsubscript{ET$\rightarrow$EN} &GEC \\
&ET$\rightarrow$EN &acc. &acc. &BLEU &BLEU &F0.5 \\\midrule
\textbf{TrTask\textsubscript{100k} + Alpacas} &50\% &59 &\textbf{76.5} &20.4 &\textbf{32.7} &56.2 \\
\textbf{TrTask\textsubscript{100k} + Alpacas} &25\% &55 &74.1 &\textbf{21.2} &32.6 &\textbf{58.1} \\
\textbf{TrTask\textsubscript{100k} + Alpacas} &0\% &56 &71.8 &21.1 &1.6 &56.2 \\\midrule
\textbf{Alpacas} &\textbf{-} &\textbf{66} &74.1 &20.8 &32.4 &50.0 \\
\bottomrule
\end{tabular}
\caption{Fine-tuning Llama-2-7B further pretrained on 1B token. Translation task ET$\rightarrow$EN direction proportion is modified. 0\% means that all of TrTask data is in EN$\rightarrow$ET direction. The amount of translation task data is fixed at 100k sentence-pairs. Results are reported on development datasets.}\label{tab:ablation-tr-direction}
\end{small}
\end{table*}

\begin{figure}
  \centering
  \begin{minipage}{0.5\textwidth}
   \newtcolorbox{alpacabox}{
  colback=white,
  colframe=black,
  arc=4pt, 
  boxrule=0.5pt, 
  fontupper=\ttfamily, 
  left=6pt, 
  right=6pt, 
}
\begin{alpacabox}
\begin{small}
Below is an instruction that describes a task, paired with an input that provides further context. Write a response that appropriately completes the request.\\
\\
\#\#\# Instruction:\\
\texttt{\{instruction\}}\\
\\
\#\#\# Input:\\
\texttt{\{input\}}\\
\\
\#\#\# Response:
\end{small}
\end{alpacabox}
  \end{minipage}
  \caption{Alpaca instruction format \citep{alpaca}}\label{tab:alpaca-format}
\end{figure}

\begin{figure}
  \centering
  \begin{minipage}{0.5\textwidth}
   \newtcolorbox{alpacabox}{
  colback=white,
  colframe=black,
  arc=4pt, 
  boxrule=0.5pt, 
  fontupper=\ttfamily, 
  left=6pt, 
  right=6pt, 
}
\begin{alpacabox}
\begin{small}
<|user|>\\
Tere!\\
<|assistant|>\\
Tere! Kas saaksin teid kuidagi aidata?\\
<|user|>\\
Kuidas alustada kirja kirjutamist?\\
<|assistant|>
\end{small}
\end{alpacabox}
  \end{minipage}
  \caption{Chat format following \citep{wang2023far}. After \texttt{<|assistant|>} the model generates the response.}\label{tab:chat-format}
\end{figure}

\begin{figure}
  \centering
  \begin{minipage}{0.5\textwidth}
   \newtcolorbox{alpacabox}{
  colback=white,
  colframe=black,
  arc=4pt, 
  boxrule=0.5pt, 
  fontupper=\ttfamily, 
  left=6pt, 
  right=6pt, 
}
\begin{alpacabox}
\begin{small}
Translate the following \{src\_lang\} text into \{tgt\_lang\}:\\
\{src\_text\}

\end{small}
\end{alpacabox}
  \end{minipage}
  \caption{Prompt used for evaluating OpenAI models on MT task.}\label{tab:openai-mt-prompt}
\end{figure}

\begin{figure*}
\centering
\begin{minipage}{\textwidth}
\newtcolorbox{alpacabox}{
  colback=white,
  colframe=black,
  arc=4pt, 
  boxrule=0.5pt, 
  left=6pt, 
  right=6pt, 
}
\begin{alpacabox}
\begin{small}
\textbf{MT evaluation prompt} \\
\texttt{Translate the following \{source language\} text into \{target language\}.} \\
\rule{\textwidth}{0.1pt} \\
\textbf{GEC evaluation prompt} \\
\texttt{Reply with a corrected version of the input sentence in \{language\} with all grammatical and spelling errors fixed. If there are no errors, reply with a copy of the original sentence.} \\
\rule{\textwidth}{0.1pt} \\
\textbf{EstCOPA evaluation prompt} (the best on development set)\\
\texttt{Olgu eelduseks lause "\{premise\}"\\
\\
Mis on selles lauses kirjeldatud olukorra \{question\}?\\
\\
VALIKUD:\\
- \{choice1\}\\
- \{choice2\}\\
}
\rule{\textwidth}{0.1pt} \\
\textbf{COPA evaluation prompt} (the best on development set)\\
\texttt{\{premise\}\\
\\
What is the \{question\} of the preceding sentence?\\
\\
OPTIONS:\\
- \{choice1\}\\
- \{choice2\}\\
}
\rule{\textwidth}{0.1pt} \\
\textbf{EstQA evaluation prompt} (the best on development set)\\
\texttt{Tekst: \{paragraph\} \\
\\
Küsimus: \{question\}\\
}
\rule{\textwidth}{0.1pt} \\
\textbf{XQuAD evaluation prompt} (the best on development set)\\
\texttt{Here is a question about this article: \{paragraph\}\\
What is the answer to this question: \{question\}}
\end{small}
\end{alpacabox}
\end{minipage}
  \caption{Instruction prompts for evaluating \textsc{LLammas}. The prompts are inserted into Alpaca-style formatting (see Table~\ref{tab:alpaca-format}). In the case of conversational model, the Alpaca-formated prompts are inserted into the chat format (see Table~\ref{tab:chat-format}).}\label{tab:eval-prompts}
\end{figure*}

\begin{table*}[h]\centering
\begin{small}
\begin{tabular}{p{12cm}}\toprule
\textbf{User:}\\
Tere! Kuidas läheb?\\\\
\textbf{Assistant:}\\
Tänan küsimast, täitsa hästi. Kuidas sinul läheb?\\\\
\textbf{User:}\\
Hästi, kuigi viimased päevad on olnud väsitavad.\\\\
\textbf{Assistant:}\\
Mis teid väsinuna tunneb? Kas teil on midagi konkreetset, mis seda põhjustab? Võite jagada oma muret ja ma aitan leida lahendusi.\\\\
\textbf{User:}\\
Keeruline öelda. Võimalik, et asi on selles, et olen liiga vähe puhanud.\\\\
\textbf{Assistant:}\\
Võib-olla peaksime proovima erinevaid stressileevendamise meetodeid nagu meditatsioon või lõdvestusharjutused. Samuti võiksime planeerida rohkem aega endale ning luua struktuurse rutiini, et tagada piisav uni ja puhkus. Räägime sellest veelgi edasi, kui see on vajalik.\\\\
\textbf{User:}\\
Paku mulle üks stressileevendusmeetod\\\\
\textbf{Assistant:}\\
Proovi regulaarselt praktiseerida joogat, mis aitab sul rahus ja harmoonias olla. Sellel on mitmeid tervistavaid efekte, sealhulgas parem kehaline painduvus, suurenenud energia ja vaimne selgus. Jooga sobib ideaalselt igapäevaselt harjumuseks muutmisena, aidates sul stressirohke elustiiliga toime tulla. Alusta tasakaaluharjutuste, venituste ja hingamisharjutustega ning avasta joogamaailmas uued võimalused endaga paremini kontaktis olemiseks. Leia sobiv stiil ja õpetaja ning alusta oma joogateekonda juba täna.\\\\
\textbf{User:}\\
Tahaks midagi lihtsamat.\\\\
\textbf{Assistant:}\\
Kirjuta endale märkmikusse positiivsed eneseütlemised, näiteks "olen tugev" või "olen nutikas". Lugege neid regulaarselt ja visualiseerige end heade omadustega inimesena. See aitab teil end motiveerida ja uskuda iseendasse. Lisage ka motivatsiooni tsitaate, nagu "Elu pole mitte ainult selle kohta, millistes asjaoludes meie juhtumeid saame, vaid ka selles, kuidas me reageerime." - Walt Disney. Tegele ka füüsiliselt aktiivsema eluga, näiteks treeningu või jalutuskäikudega värskes õhus. Need meetmed aitavad teil stressi maandada ja enda eest hoolitseda. Ärge unustage naeratada ja leida huumorit igapäevaelus. Naeratused levitavad head energiat ja loovad positiivse atmosfääri.
\\\\
\textbf{User:}\\
Tänan!\\\\
\textbf{Assistant:}\\
Olete teretulnud! Ma olen siin, et sind toetada ja aidata. Palun anna mulle teada, kui sul on mingeid muid küsimusi või abisoove. Soovin sulle imelisi hetki ja palju edu!\\
\bottomrule
\end{tabular}
\end{small}
\caption{An Example conversation with the \textsc{Llammas} model. Generated using sampling with top-p of 0.9, temperature of 0.6, repetition penalty of 1.2, and beam size of 1 (no beam search).}\label{tab:llammas-conversation-example}
\end{table*}

\end{document}